\documentclass[10pt, conference, compsocconf]{IEEEtran}
\usepackage{hyperref}
\usepackage{url}
\usepackage{amsmath,amsfonts,amssymb,epsfig}
\usepackage{graphicx}
\usepackage{caption}
\usepackage{subcaption}
\usepackage{varwidth}
\usepackage{algorithm}
\usepackage[noend]{algpseudocode}

\makeatletter
\def\BState{\State\hskip-\ALG@thistlm}
\makeatother

\title{Learning Deep Networks from Noisy Labels with Dropout Regularization}

\author{\IEEEauthorblockN{Ishan Jindal, Matthew Nokleby}
\IEEEauthorblockA{Electrical and Computer Engineering\\
Wayne State University, MI, USA\\
Email: \{ishan.jindal, matthew.nokleby\}@wayne.edu}
\and
\IEEEauthorblockN{Xuewen Chen}
\IEEEauthorblockA{Department of Computer Science\\
Wayne State University, MI, USA\\
Email: xuewen.chen@wayne.edu}
}
\begin{document}
\maketitle

\begin{abstract}

Large datasets often have unreliable labels---such as those obtained from Amazon's Mechanical Turk or social media platforms---and classifiers trained on mislabeled datasets often exhibit poor performance. We present a simple, effective technique for accounting for label noise when training deep neural networks. We augment a standard deep network with a softmax layer that models the label noise statistics. Then, we train the deep network and noise model jointly via end-to-end stochastic gradient descent on the (perhaps mislabeled) dataset. The augmented model is overdetermined, so in order to encourage the learning of a non-trivial noise model, we apply dropout regularization to the weights of the noise model during training. Numerical experiments on noisy versions of the CIFAR-10 and MNIST datasets show that the proposed dropout technique outperforms state-of-the-art methods.
\end{abstract}


\label{sect:introduction}
The previous decade has witnessed swift advances in the performance of deep neural networks for supervised image classification and recognition. State-of-the-art performance requires large datasets, such as the 10,000,000 hand-labeled images comprising the ImageNet dataset \cite{deng2009imagenet,krizhevsky2012imagenet}. Large datasets suffer from noise, not only in the images themselves, but also in their associated labels. Researchers often resort to non-expert sources such as Amazon's Mechanical Turk or tags from social networking sites to label massive datasets, resulting in unreliable labels. Furthermore, the distinction between class labels is not always precise, and even experts may disagree on the correct label of an image. Regardless its source, the resulting noise can drastically degrade learning performance \cite{zhu2004class,saez2014analyzing}.

Learning with noisy labels has been studied previously, but not extensively. Techniques for training support vector machines, $K$-nearest neighbor classifiers, and logistic regression models with label noise are presented in \cite{frenay2014classification,natarajan2013learning}. Further, \cite{natarajan2013learning} gives sample complexity bounds in the presence of label noise. Only a few papers consider deep learning with noisy labels. An early work is \cite{larsen1998design}, which studied symmetric label noise in neural networks. Binary classification with label noise was studied in \cite{mnih2012learning}. In \cite{sukhbaatar2014training}, techniques for multi-class learning and general label noise models are presented. This approach adds an extra linear layer, intended to model the label noise, to the conventional convolutional neural network (CNN) architecture. In a similar vein, the work of \cite{reed2014training} uses self-learning techniques to ``bootstrap'' the simultaneous learning of a deep network and a label noise model.

In this paper, we present a simple, effective approach to learning deep neural networks from datasets corrupted by label flips. We augment an arbitrary deep architecture with a softmax layer that characterizes the pairwise label flip probabilities. We learn jointly the parameters of the deep network and the noise model simultaneously using standard stochastic gradient descent. To ensure that the network learns an accurate noise model---instead of fitting the deep network to the noisy labels erroneously---we apply an aggressive dropout regularization to the added softmax layer. This encourages the network to learn a ``pessimistic'' noise model that denoises the corrupted labels during learning. After training, we disconnect the noise model and use the resulting deep network to classify test images. Our approach is computationally fast, completely parallelizable, and easily implemented with existing machine learning libraries \cite{jia2014caffe,collobert2011torch7,krizhevsky2012imagenet}.

%In order to avoid underfitting for deeper and deeper neural networks \cite{simonyan2014very,krizhevsky2012imagenet},a huge amount of training data is required, so the problem of label noise can be argued as the fundamental problem for image classification tasks. In this work, we have attempted to consider this problem and proposed a novel approach which modifies the simple convolutional neural network model for effective learning in the presence of label noise. This is a very simple modification, as used in \cite{sukhbaatar2014training}, where a linear layer is stacked over the final softmax output layer in order to model the noise distribution over the training classes. As a linear layer is stacked, conventional back propagation can suffice for an end to end training  and can be implemented easily with existing libraries \cite{jia2014caffe,collobert2011torch7,krizhevsky2012imagenet}. We apply an aggressive dropout regularization to the label noise transition probabilities i.e. on the weights of added linear noise layer. Main purpose of doing aggressive dropout is to learn a more pessimistic noise model. Our proposed method is very simple and easy to implement along with this our model has two main advantages. First, we do not need to normalize the linear layer weight matrix after each gradient step, this simplifies and speed up the training process. Second, the dropout is very well known technique to avoid overfitting and regularizes the weights.

In Section \ref{sect:numerical.results} we demonstrate state-of-the-art performance of the dropout-regularized noise model on noisy versions of the CIFAR-10 and MNIST datasets. In nearly all cases, the proposed method outperforms existing approaches for learning label noise models, and even for high rates of label noise. In many cases, dropout even often outperforms a genie-aided model in which the noise statistics are known {\em a priori}. We investigate the properties of the learned noise model, finding that the dropout-regularized model overestimates the label flip probabilities. We hypothesize that a pessimistic model improves performance by encouraging the deep network to cluster images naturally when confronted with conflicting image labels.

%We demonstrate the image classifcation accuracy of proposed method on deep CNNs and simple deep networks for CIFAR-10 and MNIST dataset, where we synthetically corrupt the image labels. In the experiments, we also study how the different kind of noise models effect the classification accuracy of proposed algorithm. We  compare the proposed method with the state of the art methods and claims better classification accuracy. In some cases the proposed method works better than the standard classification accuracy rate, when the true noise model is known.  We believe that this is because the learned noise model is "pessimistic``, overestimating the model noise, which encourages a more noise-aware learning of the neural network.

%The remainder of this paper is structured as follows. In Section \ref{sect:system.model} we define the noisy learning problem formally. In Section \ref{sect:dropout} we detail the proposed dropout-regularized learning approach.  Finally, in Section \ref{sect:conclusion} we draw conclusions.
%\input{related.work.tex}
 
\section{Problem Statement}\label{sect:system.model}
In the usual supervised learning setting, we have access to a set of $n$ labeled training images. Denote each image by $x_i \in \mathbb{R}^d$, and denote the class label for image $x_i$ by $y_i~\in~\{1,\dots,C\}$. Denote this ideal training set by
$$ \mathcal{D} = \{(x_1,y_1),(x_2,y_2),...,(x_n,y_n)\}.$$
As discussed in the introduction, accurate labels are difficult to obtain for large datasets, so we suppose that we have access only to noisy labels, denoted by $y_i^\prime$. Denote the noisy training set by
$$ \mathcal{D}^\prime = \{(x_1,y^\prime_1),(x_2,y^\prime_2),...,(x_n,y^\prime_n)\}.$$

We assume a probabilistic model of label noise in which each noisy label $y^\prime$ depends only on the true label $y$ and not on the image $x$. We further suppose that the noisy labels are i.i.d. conditioned on the true labels. That is, $y^\prime_i$ and $y_j^\prime$ are independent of each other given the true labels $y_i$ and $y_j$, and $p(y^\prime_i|y_i) = p(y^\prime_j|y_j)$ for image pairs $x_i$ and $x_j$. We represent the conditional noise model by the column-stochastic matrix $\Psi \in \mathbb{R}^{C \times C}$:
\begin{equation}\label{eqn:true.noise.model}
	p(y^\prime=i | y = j) = \Psi_{ij},
\end{equation}
where $\Psi_{ij}$ is the $(i,j)$th element of $\Psi$.

In our simulations, we synthesize the noisy labels. From the standard datasets CIFAR-10 and MNIST, we fix a noise distribution $\Psi$ and create noisy labels by drawing i.i.d. from the distribution specified by (\ref{eqn:true.noise.model}) for the training samples. We do not perturb the labels for the test samples.

While the proposed method works for any $\Psi$, we use two parametric noise models in the sequel. First, we choose a noise level $p$, and we set
\begin{equation}\label{eqn:uniform.noise}
	\Psi = (1-p)\mathbf{I} + \frac{p}{C}\mathbf{1}\mathbf{1}^T,
\end{equation}
where $\mathbf{I}$ is the identity matrix and $\mathbf{1}$ is the all-ones column vector. That is, the noisy label is the true label with probability $1-p$ and is drawn uniformly from $\{1,\dots,C\}$ with probability p. We call this the {\em uniform noise} model.

Second, we again choose a noise level $p$, and we set
\begin{equation}\label{eqn:non.uniform.noise}
	\Psi = (1-p)\mathbf{I} + p\Delta,
\end{equation}
where the columns of $\Delta$ are drawn uniformly from the unit simplex, i.e. the set of vectors with nonnegative elements that sum to one. The matrix $\Delta$ is constant over a single instantiation of the noisy training set $\mathcal{D}^\prime$. We call this the {\em non-uniform} noise model.

\subsection{Learning Deep Networks with Noise Models}
\begin{figure*}[t]
        \centering
        \begin{subfigure}[b]{0.5\textwidth}
                \centering
                \includegraphics[width=\textwidth]{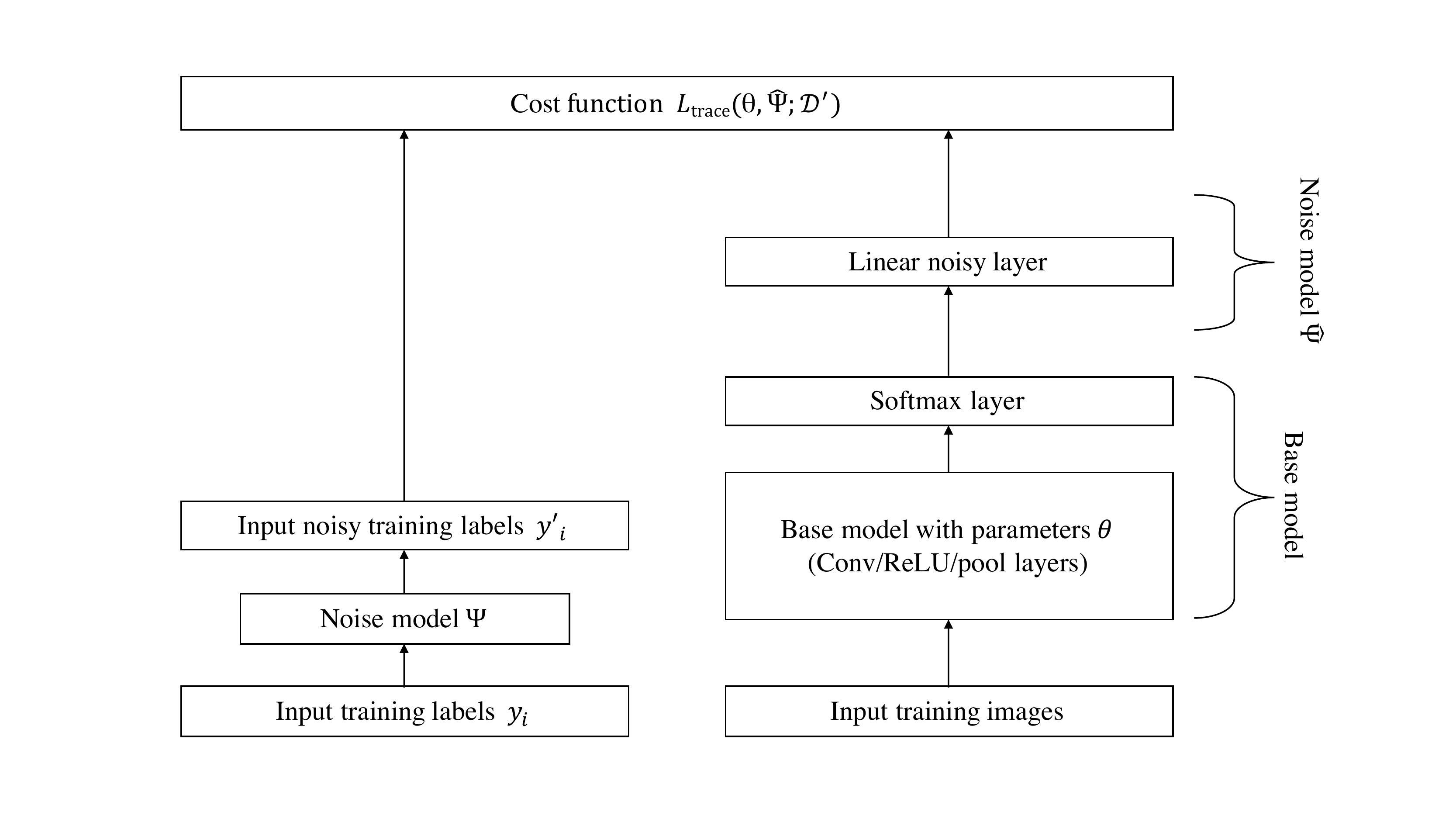}
                \caption{A deep network augmented with a linear noise model.}
                \label{linear.softmax}
        \end{subfigure}%
        ~
        \begin{subfigure}[b]{0.5\textwidth}
                \centering
                \includegraphics[width=\textwidth]{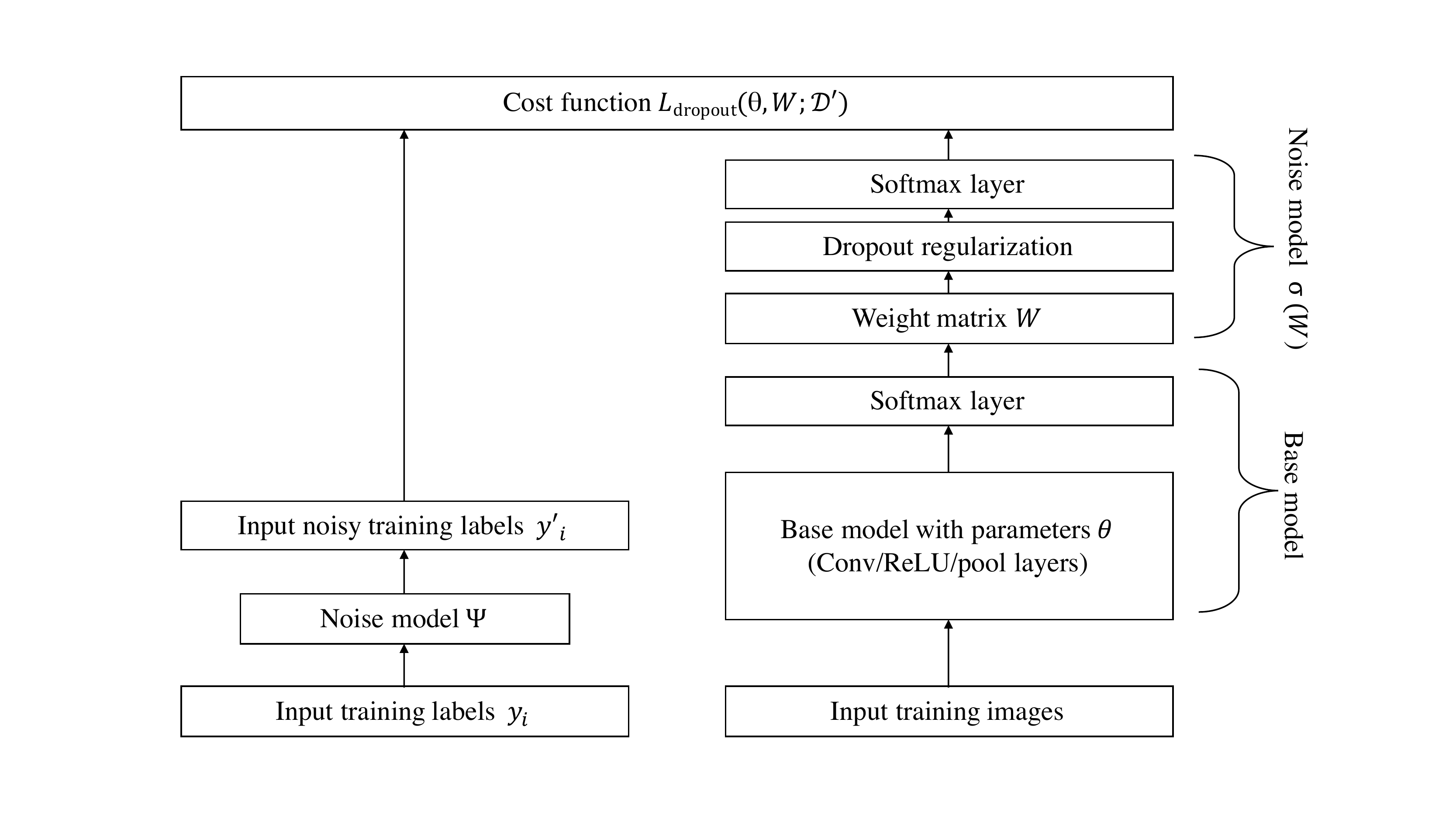}
                 \caption{A deep network augmented with a softmax/dropout noise model.}
                \label{fig:softmax.noise}
        \end{subfigure}%
        \caption{}\label{fig:linear.noise}
\end{figure*}

Our objective is to learn a deep network from the noisy training set $\mathcal{D}^\prime$ that accurately classifies cleanly-labeled images. Our approach is to take a standard deep network---which we call the {\em base model}---and augment it with a noise model that accounts for label noise. Then, the base and noise models are learned jointly via stochastic gradient descent. The noise model has a role only during training---as the noise model is learned, it effectively denoises the labels during backpropagation, making it possible to learn a more accurate base model. After training, the noise model is disconnected, and test images are classified using the base model output.

We use two standard deep networks for the base model. The first is the %[``AlexNet''] 
deep convolutional network. It has three processing layers, with rectified linear units (ReLus) and max- and average-pool operations between layers. The hyperparameters are similar to those used in the popular ``AlexNet'' architecture, described in \cite{krizhevsky2012imagenet}. The second model is a standard deep neural network, with three rectified linear processing layers (RELUs). %The hyperparameters---the number of units per layer, etc.---depend on which training set is used, and we specify these in describing the numerical results in Section \ref{sect:numerical.results}.

We lump the base model parameters---processing layer weights and biases, etc.---into a single parameter vector $\theta$. Further, let $h$ be the output vector of the final layer of the base model. Define the usual softmax function
\begin{equation}
	\sigma(\mathbf{x})_i = \frac{\exp(x_i)}{\sum_j \exp(x_j)}.
\end{equation}
Then, for test image $x$, the base model estimate of the distribution of the class label is
\begin{equation}
	p(\hat{y} | x; \theta) = \sigma(h).
\end{equation}

One approach to noisy labels is to use the base model without modification and treat $y_i^\prime$ as the true label for $x_i$. Taking the standard cross-entropy loss, one can minimize the empirical risk
%\begin{align}\label{eqn:base.model.loss}
%      L_\mathrm{base}(\theta;\mathcal{D}^\prime) &= -\frac{1}{N}\sum_{i=1}^n \sum_{c=1}^C \#(y_i^\prime = c)\log(p(\hat{y} = c | x_i; \theta)) \\
%      &= -\frac{1}{N}\sum_{i=1}^n \sum_{c=1}^C \#(y_i^\prime = c)\log(\sigma(h)_c),
%\end{align}
\begin{align}\label{eqn:base.model.loss}
      L_\mathrm{base}(\theta;\mathcal{D}^\prime) &= -\frac{1}{n}\sum_{i=1}^n \log(p(\hat{y} = y^\prime_i | x_i; \theta)) \\
      &= -\frac{1}{n}\sum_{i=1}^n \log(\sigma(h)_{y^\prime_i}),
\end{align}
As shown in Section \ref{sect:numerical.results}, the base model alone offers satisfactory performance when the label noise is not too severe; otherwise the incorrect labels overwhelm the model, and it fails. %Overall, the performance is substantially improved by including an explicit noise model.

To motivate our approach, we describe first the method presented in \cite{sukhbaatar2014training}. Suppose momentarily that the true noise distribution, characterized by $\Psi$, is known. One can augment the base model with a linear noise model, with weight matrix equal to $\Psi$, as depicted in Figure \ref{linear.softmax}. For this architecture, we can express the estimate of the distribution of the {\em noisy} class label as
\begin{align}
	p(\hat{y}^\prime | x; \theta,\Psi) &= \sum_{c=1}^C p(\hat{y}^\prime | \hat{y} = c)p(\hat{y} = c) \\
	&= \Psi \cdot \sigma(h),
\end{align}
where $\cdot$ is standard matrix-vector multiplication. We can then minimize the empirical cross-entropy of the noisy labels directly:
%\begin{align}\label{eqn:true.noise.loss}
%      L_\mathrm{kn}(\theta;\mathcal{D}^\prime,\Psi) &= -\frac{1}{N}\sum_{i=1}^n \sum_{c=1}^C \#(y_i^\prime = c)\log(p(\hat{y}^\prime = c | x_i; \theta)) \\
%      &= -\frac{1}{N}\sum_{i=1}^n \sum_{c=1}^C \#(y_i^\prime = c)\log([\Psi \cdot \sigma(h)]_c),
%\end{align}
\begin{align}\label{eqn:true.noise.loss}
      L_\mathrm{true}(\theta;\mathcal{D}^\prime,\Psi) &= -\frac{1}{n}\sum_{i=1}^n \log(p(\hat{y}^\prime = y^\prime_i | x_i; \theta)) \\
      &= -\frac{1}{n}\sum_{i=1}^n\log([\Psi \cdot \sigma(h)]_{y^\prime_i}),
\end{align}
where $[\cdot]_i$ returns the $i$th element of a vector. Then, each test sample $x$ is classified according to the output of the base model, i.e. $\sigma(h)$. Because the noise model is known perfectly, one might expect that this approach gives the best possible performance. While it does provide excellent performance, in Section \ref{sect:numerical.results} we show that even better performance is possible in most cases.

The noise model, however, is usually unknown. Furthermore, we do not know which labels are corrupted and we cannot estimate a noise model directly. The authors of \cite{sukhbaatar2014training} suggested that one can estimate the noise probabilities $\Psi$ while simultaneously learning the base model parameters $\theta$. The challenge here is that convolutional networks are sufficiently expressive models that base model may fit to the noisy labels directly and learn a trivial noise model. To prevent this, the authors of \cite{sukhbaatar2014training} add a regularization term that penalizes the trace of the estimate of $\Psi$. This encourages a diffuse noise model estimate and permits the base model to learn from denoised labels. The associated loss function is
%\begin{align}\label{eqn:trace.loss}
%      L_\mathrm{tr}(\theta,\hat{\Psi};\mathcal{D}^\prime) &= -\frac{1}{N}\sum_{i=1}^n \sum_{c=1}^C \#(y_i^\prime = c)\log(p(\hat{y}^\prime = c | x_i; \theta)) + \lambda \mathrm{tr}(\Psi) \\
%      &= -\frac{1}{N}\sum_{i=1}^n \sum_{c=1}^C \#(y_i^\prime = c)\log([\hat{\Psi} \cdot \sigma(h)]_c) + \lambda \mathrm{tr}(\Psi),
%\end{align}

\begin{align}\label{eqn:trace.loss}
      L_\mathrm{trace}(\theta,\hat{\Psi};\mathcal{D}^\prime) &= -\frac{1}{n}\sum_{i=1}^n \log(p(\hat{y}^\prime = y^\prime_i | x_i; \theta)) + \lambda \mathrm{tr}(\Psi) \\
      &= -\frac{1}{n}\sum_{i=1}^n \log([\hat{\Psi} \cdot \sigma(h)]_{y^\prime_i}) + \lambda \mathrm{tr}(\Psi),
\end{align}
where $\mathrm{tr}(\cdot)$ is the matrix trace, and $\lambda$ is a regularization parameter chosen via cross-validation. When minimizing $L_\mathrm{trace}$, one must take care to project the estimate $\hat{\Psi}$ onto the space of stochastic matrices at every iteration, else it will not correspond to a meaningful model of label noise.

%This approach also gives good performance, although the performance is worse than in the case of known $\Psi$.

\section{Dropout Regularization}\label{sect:dropout}
We propose to augment the base model with a different noise architecture. As depicted in Figure \ref{fig:softmax.noise}, we add a softmax layer with square weight matrix $W \in \mathbb{R}^{C \times C}$, unconstrained. We interpret the output of this softmax layer, denoted $g = \sigma(W h)$, as the probability distribution over the noisy label $y^\prime$.
%To the output of the base model softmax layer we add a linear layer as before, after which we append a softmax layer. We take the output of the base model softmax layer, denoted as $\mathbf{h} \in \mathbb{R}^{|\mathcal{C}|}$, as a categorical probability distribution over the true labels $y_i$. Denote the weights of the linear noise layer as $\mathbf{W} \in \mathbb{R}^{|\mathcal{C}| \times |\mathcal{C}|}$. Then, we take the output of the final softmax layer as a categorical probability distribution over the noisy labels $y_i^\prime$.
This results in the effective conditional probability distribution of the noisy label $y^\prime$ conditioned on $y$:
\begin{equation}\label{eqn:softmax.noise.model}
		p(y^\prime=i | y = j) = [\sigma(W e_j)]_i,
		%\frac{\exp\left(W_{ij}\right)}{\sum_{l=1}^{|\mathcal{C}|}\exp\left(W_{lj} \right)},
\end{equation}
where $e_j$ is the $j$th elementary vector. We use this architecture without loss of generality. Because the softmax function is invertible, there is a one-to-one relationship between noise distributions induced by $\Psi$ and (\ref{eqn:true.noise.model}) and those induced by $W$ and (\ref{eqn:softmax.noise.model}). %For any $\Psi$, there exists a $W$ that results in the same conditional distribution on the labels.
For any $W$ and base model parameters $\theta$, the estimate of the distribution of the noisy class label is
\begin{align}
	p(\hat{y}^\prime | \hat{y}) &= \sum_{c=1}^C p(\hat{y}^\prime | \hat{y} = c)p(\hat{y} = c) \\
	&= \sigma(W \cdot \sigma(h)).
\end{align} 

This architecture offers two major advantages. First, the matrix $W$ is unconstrained during optimization. Because the softmax layer implicitly normalizes the resulting conditional probabilities, there is no need to normalize $W$ or force its entries to be nonnegative. This simplifies the optimization process by eliminating the normalization step described above.

Second, it is congruent with dropout regularization, which we apply to the output of base model $\sigma(h)$  to prevent the base model from learning the noisy labels directly. Dropout is a well-established technique for preventing overfitting in deep learning \cite{srivastava2014dropout}. It regularizes learning by introducing binary multiplicative noise during training. At each gradient step, the base model outputs are multiplied by random variables drawn i.i.d from the Bernoulli distribution $\mathrm{Bern}(q)$. This ``thins'' out the network, effectively sampling from a different network for each gradient step.

Applying dropout to $\sigma(h)$ entails forming the effective weight matrix

\begin{align}
	a & \sim \mathrm{Bern}(q)\\
	\widehat{\sigma(h)} & = a \odot \sigma(h)
\end{align}

where $a$ has entries drawn i.i.d. from the Bernoulli distribution $\mathrm{Bern}(q)$ and $\odot$ represents the Hadamard (element-wise) product. We choose a different vector $a$ for each mini-batch, i.e. each SGD step, in the training set. Again using the cross-entropy loss, the resulting loss function is
%\begin{align}
%      L_\mathrm{dr}(\theta,W;\mathcal{D}^\prime) &= -\frac{1}{N}\sum_{i=1}^n \sum_{c=1}^C \#(y_i^\prime = c)\log(p(\hat{y}^\prime = c | x_i; \theta)) \\
%      &= -\frac{1}{N}\sum_{i=1}^n \sum_{c=1}^C \#(y_i^\prime = c)\log([\sigma((W\odot A_n) \cdot \sigma(h))]_c).
%\end{align}

\begin{align}
      L_\mathrm{dropout}(\theta,W;\mathcal{D}^\prime) &= -\frac{1}{n}\sum_{i=1}^n \log(p(\hat{y}^\prime = y^\prime_i | x_i; \theta)) \\
            &= -\frac{1}{n}\sum_{i=1}^n \log([\sigma(W \cdot (a \odot \sigma(h))]_{y^\prime_i})
\end{align}

Observing the conditional distribution in (\ref{eqn:softmax.noise.model}), each instantiation of the multiplicative noise $a$ zeros out a fraction of the elements $W_{ij}$, forcing the associated probabilities to a baseline, uniform value. [ISHAN: Is this right?] This forces the learning ``action'' on the remaining probabilities, which encourages a non-trivial noise model. The Bernoulli parameter $q$ determines the sparsity of each instantiation. In our simulations, we find that $q=0.1$---which corresponds to an aggressively sparse model---works best.

The usual dropout procedure involves ``averaging'' together the different models when classifying samples by reducing the learned weights. In our setting, this is unnecessary. The noise model serves only as an intermediate step for denoising the noisy labels to train a more accurate base model. The noise model is disconnected at test time, and averaging is not performed.

\section{Experimental Results}\label{sect:numerical.results}
In this section, we demonstrate the performance of the proposed method. We state results on two datasets (CIFAR-10 and MNIST), two noise models (uniform and non-uniform), and two base models (CNN and DNN). For training the CNN, we use the model architecture from the publicly-available MATLAB toolbox $\mathtt{MathConvNet}$ \cite{vedaldi2015matconvnet}. [ISHAN: What are the hyperparameters of this model?] Other than changing the size of the input units, we keep the model hyperparameters constant. For training the DNN, we use the architecture used in \cite{reed2014training}, which has $784-500-300-10$ ReLUs per layer. In each case, we present results for label noise probabilities $p \in \{0.3,0.5,0.7\}$, i.e. label noise that corrupts 30\%, 50\%, and 70\% of the training samples. As mentioned earlier, we use a dropout rate of $q=0.1$ in all simulations. We train the CNN and DNN end-to-end using stochastic gradient descent with batch size 100. When training on the MNIST dataset, we perform early stopping, ceasing iterations when the loss function begins to increase. We emphasize that the loss function does not depend on the true labels, so choosing when to stop does not require knowledge of the uncorrupted dataset. MATLAB code for these simulations is available at \cite{CODE}.

\subsection{CIFAR Images}
The CIFAR-10 dataset \cite{krizhevsky2009learning} is a subset of the Tiny Images dataset \cite{torralba200880}. CIFAR-10 consists of 50,000 training images and 10,000 test images, each of which belongs to one of ten object categories, which are equally represented in the training and test sets. Each image has dimension $32 \times 32 \times 3$, where the latter dimension reflects the three color channels of the images.

First, we state results for the uniform noise model using CNN. For $p \in \{0.3,0.5,0.7\}$, we choose $\Psi~=~(1-p)\mathbf{I}~+~p/C \mathbf{11}^T$ as indicated in (\ref{eqn:uniform.noise}). We corrupt the labels in the CIFAR-10 training according to $\Psi$, and we leave the test labels uncorrupted. For reference, CNN achieves 20.49\% classification error when trained on the noise-free dataset.

We state the classification accuracy over the test set in Table \ref{cifuni}. As a baseline, we present results for the {\em base model}, in which the noisy labels are treated as true labels and the model parameters are chosen to minimize the standard loss function in (\ref{eqn:base.model.loss}). We also present results for the {\em true noise} model, in which $\Psi$ is known, a linear noise layer with weights $\Psi$ is appended to the base model, and the model parameters are chosen to minimize the loss function in (\ref{eqn:base.model.loss}). Next, we present results for the proposed softmax architecture, first without regularization (referred to as ``Softmax'' in Table \ref{cifuni}) and then with the proposed dropout regularization (``Dropout'').

Finally, we compare to the results presented in \cite{sukhbaatar2014training} (``Trace''), in which a linear layer is added, but the label noise model $\Psi$ is learned jointly with the base model parameters according to the trace-penalized loss function of (\ref{eqn:trace.loss}). We emphasize that these results come with significant caveats. While the noise level and network architecture used here is the same as that of \cite{sukhbaatar2014training}, the authors of \cite{sukhbaatar2014training} used a non-uniform noise model which we do not replicate in this paper. Therefore, these results are from a roughly comparable, but not strictly identical, noise scenario.

%Here, the base model is defined as the raw model without noise modeling, i.e. no extra linear noise layer is added to the conventional CNN model and all the noise is learned by the CNN itself. For all the experiments no noise is added to the test samples and remained intact. The learned noise model is the one where the linear noise layer is learned by the proposed approach and the true noise model is when the learned noisy layer is replaced by the known noisy distribution. \\
%In table \ref{cifuni} results are shown with 50K training images and varying noise levels and uniform nosie distribution. We have used the same setting for learning linear noise layer as described by \cite{sukhbaatar2014training}. For our CNN model when all the labels are correct the calculated error rate is 20.49\%.
%From the table \ref{cifuni}, it can be observed that the learned noise distribution learned by proposed approach can withstand up to 50\% of the noise in comparison with the another method. Even at 70\% of the noise our proposed method shows 51.6\% of classification accuracy whereas other method didn't even shows results with 70\% noise. 

\begin{table*}[t]
\centering
\caption{Classification accuracy on the CIFAR-10 dataset with uniform label noise and the CNN architecture.}
\label{cifuni}
\begin{tabular}{|c|c|c|c|c|c|}
\hline
Noise level & True noise & Base model & Softmax & Dropout & Trace (\cite{sukhbaatar2014training}) \\ \hline
30\%        & 25.76                                                       & 29.78 & 26.04    & 24.43           & 26                                                            \\ \hline
50\%        & 29.63                                                       & 38.76 & 33.40    & 32.64           & 35                                                            \\ \hline
70\%        & 36.24                                                       & 48.34 & 37.10    &   33.00          & 63                                                            \\ \hline
\end{tabular}
\end{table*}

In most cases, the proposed dropout method gives the best performance---even better than the true noise model, which supposes that $\Psi$ is known {\em a priori}. Only in the case of 50\% noise does the true noise model outperform dropout. Note that even without dropout regularization, the proposed softmax noise model gives satisfactory performance, consistently outperforming the base model.

\begin{figure*}[t]
        \centering
        \begin{subfigure}[b]{0.3\textwidth}
                \centering
                \includegraphics[width=\textwidth]{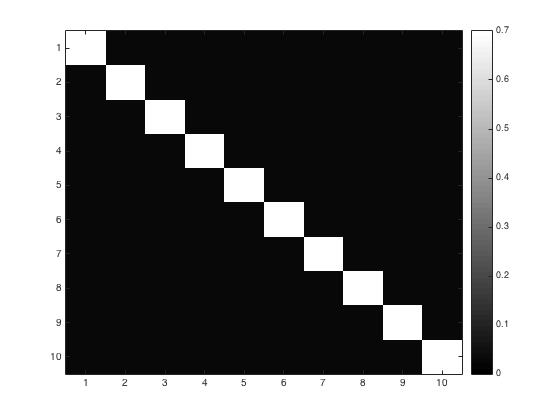}
                \caption{30\% True Noise}
                \label{noi1Uni}
        \end{subfigure}%
        \quad
        \begin{subfigure}[b]{0.3\textwidth}
                \centering
                \includegraphics[width=\textwidth]{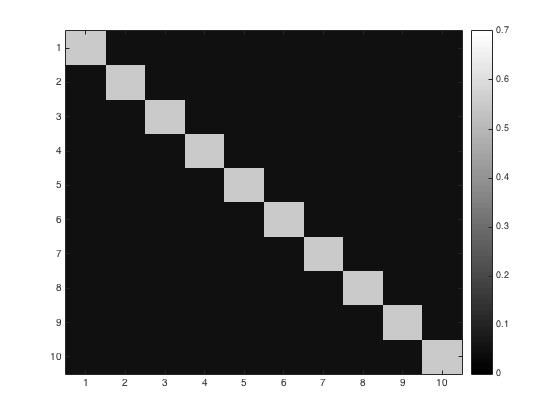}
                \caption{50\% True Noise}
                \label{noi2Uni}
        \end{subfigure}%
        \quad
        \begin{subfigure}[b]{0.3\textwidth}
                \centering
                \includegraphics[width=\textwidth]{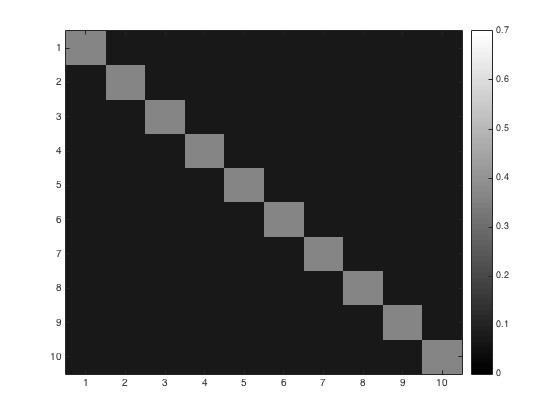}
                \caption{70\% True Noise}
                \label{noi3Uni}
        \end{subfigure}%
        
                \begin{subfigure}[b]{0.3\textwidth}
                \centering
                \includegraphics[width=\textwidth]{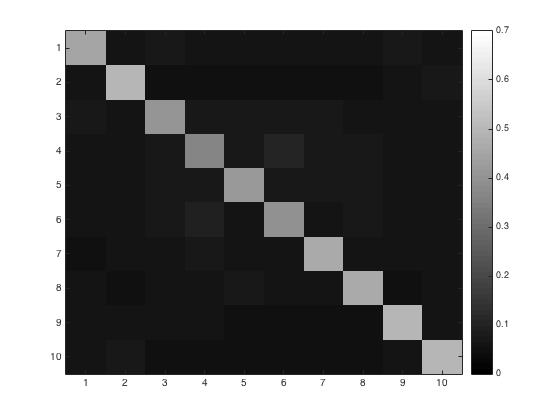}
                \caption{30\% Learned Noise}
                \label{len1Uni}
        \end{subfigure}%
        \quad
        \begin{subfigure}[b]{0.3\textwidth}
                \centering
                \includegraphics[width=\textwidth]{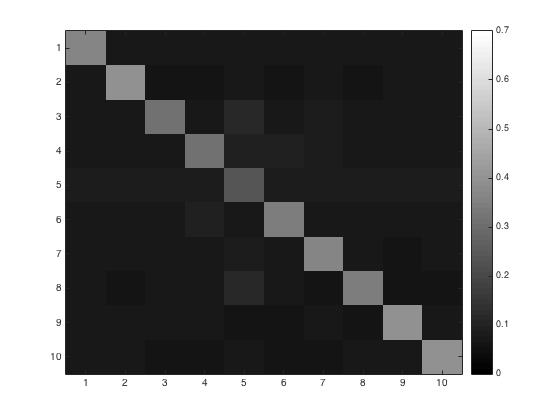}
                \caption{50\% Learned Noise}
                \label{len2Uni}
        \end{subfigure}%
        \quad
        \begin{subfigure}[b]{0.3\textwidth}
                \centering
                \includegraphics[width=\textwidth]{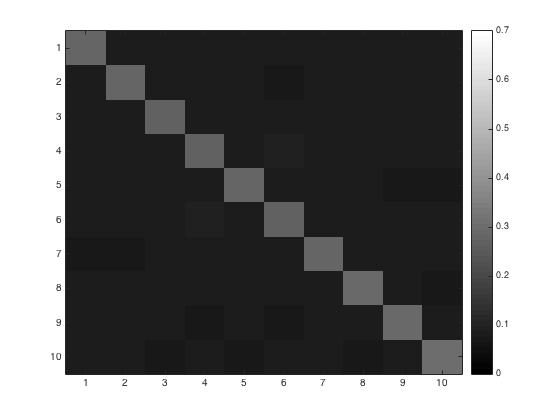}
                \caption{70\% Learned Noise}
                \label{len3Uni}
        \end{subfigure}%
        \caption{True and learned uniform noise distributions. The first row shows the elements of the true noise matrix $\Psi$ for the uniform noise model with 30\%, 50\% and 70\% noise levels. The second row shows the noise model learned via the proposed dropout method.}\label{noise_model_Uniform}
\end{figure*}

Because there is a one-to-one relationship between the softmax and linear noise models, one might expect their performance to be similar. To understand further why this is not so, in Figure \ref{noise_model_Uniform} we plot the true noise model $\Psi$ alongside the equivalent noise matrices learned via the proposed dropout scheme. The learned models are of the correct form---approximately uniform and diagonally dominant---but they also are more pessimistic, underestimating the probability of a correct noise label by a few percent. Indeed, the average diagonal value of the learned noise matrices are $0.279, 0.345,$ and $0.447$ for 30\%, 50\%, and 70\% noise, respectively. This suggests that a CNN may learn from noisy labels better if the denoising model is pessimistic. This notion is a topic for future investigation.

Next, we state results for the non-uniform noise model using a CNN. For $p \in \{0.3,0.5,0.7\}$, we corrupt the labels in the CIFAR-10 training set according to  $\Psi~=~(1-p)\mathbf{I}~+~p\Delta$ as indicated in (\ref{eqn:non.uniform.noise}). We again compare the proposed dropout scheme to the base model, the true noise model, and the trace-regularized scheme of \cite{sukhbaatar2014training}. We emphasize again that these error rates, taken directly from \cite{sukhbaatar2014training}, are for a similar but not identical noise model. We omit results for the unregularized softmax scheme.

\begin{figure*}[t]
        \centering
        \begin{subfigure}[b]{0.3\textwidth}
                \centering
                \includegraphics[width=\textwidth]{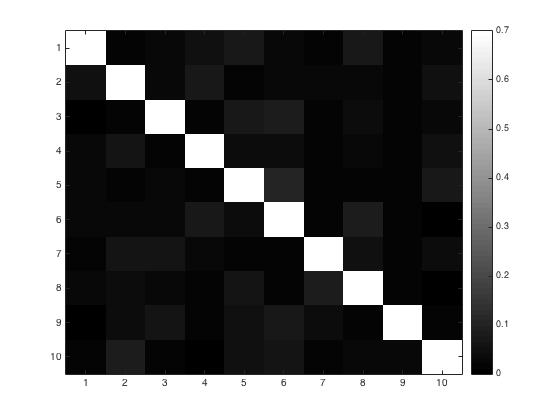}
                \caption{30\% True Noise}
                \label{noi1}
        \end{subfigure}%
        \quad
        \begin{subfigure}[b]{0.3\textwidth}
                \centering
                \includegraphics[width=\textwidth]{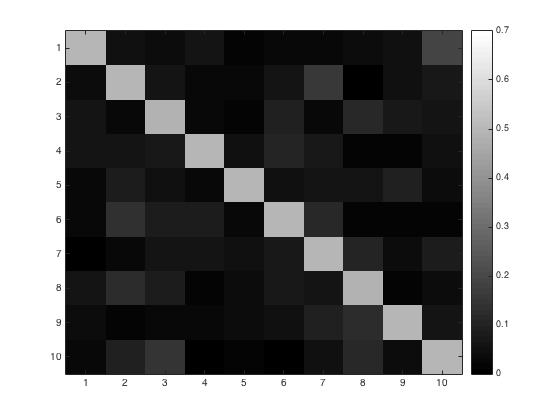}
                \caption{50\% True Noise}
                \label{noi2}
        \end{subfigure}%
        \quad
        \begin{subfigure}[b]{0.3\textwidth}
                \centering
                \includegraphics[width=\textwidth]{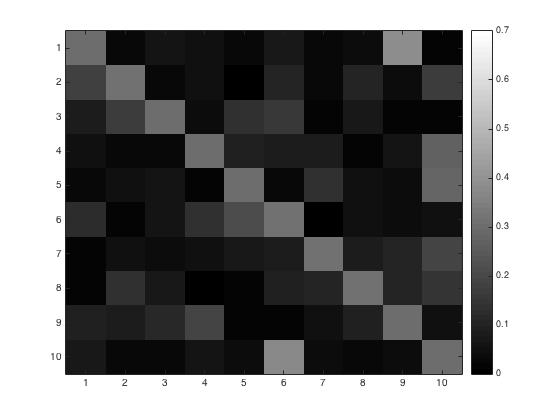}
                \caption{70\% True Noise}
                \label{noi3}
        \end{subfigure}%
        
                \begin{subfigure}[b]{0.3\textwidth}
                \centering
                \includegraphics[width=\textwidth]{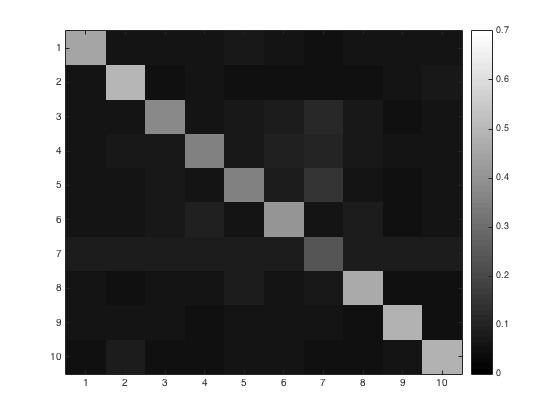}
                \caption{30\% Learned Noise}
                \label{len1}
        \end{subfigure}%
        \quad
        \begin{subfigure}[b]{0.3\textwidth}
                \centering
                \includegraphics[width=\textwidth]{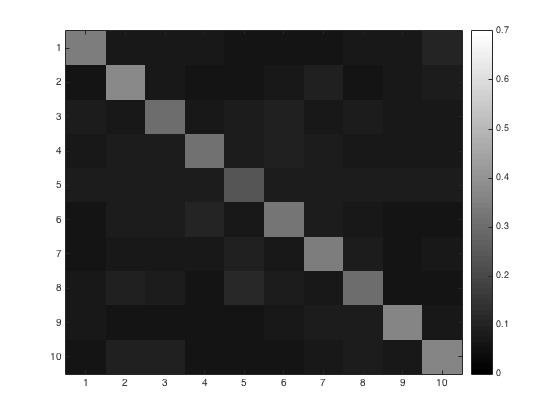}
                \caption{50\% Learned Noise}
                \label{len2}
        \end{subfigure}%
        \quad
        \begin{subfigure}[b]{0.3\textwidth}
                \centering
                \includegraphics[width=\textwidth]{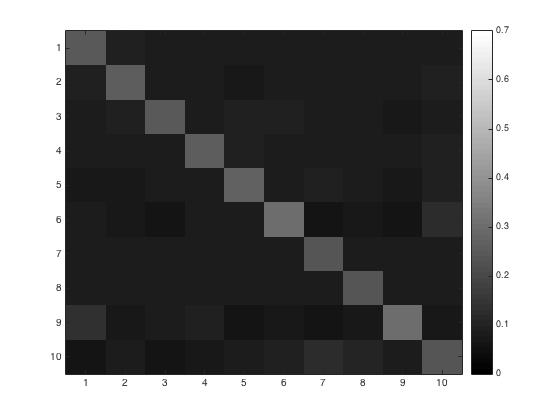}
                \caption{70\% Learned Noise}
                \label{len3}
        \end{subfigure}%
        \caption{True and learned non-uniform noise distributions. The first row shows the elements of the true noise matrix $\Psi$ for the non-uniform noise model with 30\%, 50\% and 70\% noise levels. The second row shows the noise model learned via the proposed dropout method. }\label{noise_model}
\end{figure*}

Table \ref{cifrand} states the classification error for the different schemes over the CIFAR-10 test set. Again dropout performs well, outperforming the base model and performing better or on par with the trace-regularized scheme. In this case, however, dropout does not outperform the true noise model. Indeed, overall dropout performs worse under non-uniform noise. To investigate this further, we plot the values of $\Psi$ used for simulations and the noise model learned via dropout in Figure \ref{noise_model}. Similar to before, dropout learns a more pessimistic noise model, with average diagonal entries equal to $0.256$, $0.326$, and $0.4125$ for 30\%, 50\%, and 70\% noise levels, respectively.  Further, the learned noise models are close to uniform, even though the true model is non-uniform. We hypothesize that the failure of dropout to learn a non-uniform noise model explains the performance gap. We emphasize, though, the state-of-the-art performance of the model learned by dropout. %As in the uniform case, the role of noise model fit in the training of CNNs is a topic for further investigation.

\begin{table}[htb]
\centering
\caption{Classification error rates on the CIFAR-10 dataset with non-uniform label noise and the CNN architecture.}
\label{cifrand}
\begin{tabular}{|c|c|c|c|c|}
\hline
Noise level & True noise & Base model & Dropout  & Trace (\cite{sukhbaatar2014training}) \\ \hline
30\%        & 24.95                                                       & 30.49      & 25.4            & 26                                                            \\ \hline
50\%        & 29.9                                                        & 39.47      & 31.28           & 35                                                            \\ \hline
70\%        & 63.91                                                       & 65.6       & 63.04           & 63                                                            \\ \hline
\end{tabular}
\end{table}

\subsection{MNIST Images}
MNIST is a set of images of handwritten digits \cite{lecun1998mnist}. It has 60,000 training images and 10,000 test images. We use the version of the dataset included in $\mathtt{MatConvNet}$, in which the original black-and-white images are normalized to grayscale and fit to a dimension of $28 \times 28$. For reference, the CNN achieves 0.89\% classification error when trained on the uncorrupted training set.

First, we present results for learning the CNN model parameters on the MNIST training set corrupted by uniform noise. As usual we take $\Psi$ as defined in (\ref{eqn:uniform.noise}) for $p \in \{0.3,0.5,0.7\}$. We compare the proposed dropout method to the base and true noise models. For this scenario, there is no prior work against which to compare.

\begin{table}[htb]
\centering
\caption{Classification error rates for the CNN architecture trained on the MNIST dataset corrupted by uniform noise.}
\label{mniunicnn}
\begin{tabular}{|c|c|c|c|c|}
\hline
Noise level & True noise & Base model & Dropout \\ \hline
30\%        & 1.3                                                        & 8.3        & 1.2                                                                         \\ \hline
50\%        & 2.06                                                       & 25.44      & 1.92            \\ \hline
70\%        & 3.31                                                       & 44.42      & 3.12           \\ \hline
\end{tabular}
\end{table}

We state the results in Table \ref{mniunicnn}. Dropout outperforms the true noise model for 30\% and 50\% noise, and performs only slightly worse at 70\% noise. Still, dropout proves quite robust to label noise, outperforming the base model substantially.

%MNIST is handwritten digit dataset, having 60k training images and 10k test images. In the same setting only the training image labels are introduces to noise and test image labels are not distrubed. For this dataset we performed the same kind of experiments as in the case of CIFAR-10 dataset with varying degree of noise in training labels and both with random and uniform noise. Similarly, We have obtained results for the base model and the learned noise distributions model as well. When no noise in labels is introduced the used CNN model has 0.89\% error rate. 

%Table\ref{mniunicnn} shows result of proposed method when the base model is CNN with uniform noise and compares it with the base model and true noise model. Same model is studied with random noise distribution in table \ref{mnirandcnn}. In uniform noise distribution our proposed model works better than the true noise model, upto 50\% of incorrect labels, as seen similarly in CIFAR 10 dataset results.

In Table \ref{mnirandcnn} we state the results of the same experiment, this time with $\Psi$ drawn according to the non-uniform noise model of (\ref{eqn:non.uniform.noise}). Similar to the CIFAR-10 case, the relative performance of dropout is worse. It slightly under-performs relative to the true noise model for 30\% and 50\%, and it performs substantially worse for 70\%. This is due to two factors: first, the dropout scheme learns non-uniform noise models poorly, as seen above, and the MNIST dataset does not cluster as naturally as the CIFAR-10 dataset. %Images of different digits are more alike than images of different objects, and noisy labels make it difficult to learn a powerful classifier.

\begin{table}[h]
\centering
\caption{Classification error rates for the CNN architecture trained on the MNIST dataset corrupted by non-uniform noise.}
\label{mnirandcnn}
\begin{tabular}{|c|c|c|c|}
\hline
Noise Level & \begin{tabular}[c]{@{}c@{}}True Noise\\ Model\end{tabular} & Base model & Dropout  \\ \hline
30\%        & 1.72                                                      & 4.5        & 1.83            \\ \hline
50\%        & 2.29                                                       & 34.5       & 2.83           \\ \hline
70\%        & 3.58                                                       & 48.80      & 24.6            \\ \hline
\end{tabular}
\end{table}
%Here, In order to have exact comparison of results, we implemented a deep neural network architecture as described in \cite{reed2014training}, having $784-500-300-10$ neural network with rectified linear units and trained the network with mini-batch SGD. For no noise this model results in 1.84\% classification error.

%Results are shown in table \ref{mniunideep} and compared with \cite{reed2014training}. 
%We are different form their method in the sense that our base model is Convolutional neural neywork  whereas \cite{reed2014training} uses the simple deep neural networks. It can be well noticed from the results that the base model in our approach performs quite good then the results shown in \cite{reed2014training}. In order to have proper comparison we have computed the classification accuracy with 48\% and 50\%  noise distribution and a good deal of accuracy drop is observed as we move to 50\% noisy labels from 48\%. Our proposed method performs quite good when the base model is CNN and the linear noise layer is diffused according to (\ref{eqlambda}).

To compare the dropout performance on MNIST with previous work, we also state results for a three-layer DNN as described in \cite{reed2014training}. As mentioned above, this network has $784-500-300-10$ rectified linear units per layer. The DNN is less sophisticated than the CNN, so it has worse performance overall. When trained on the uncorrupted MNIST training set, it achieves 1.84\% classification error.

We first state results for uniform noise, shown in Table \ref{mniunideep}. As before, we corrupt the MNIST training set labels with noise drawn according to (\ref{eqn:uniform.noise}). In addition to the true noise and base models, we compare the proposed dropout scheme to that presented in \cite{reed2014training}, where a ``bootstrapping'' scheme is used to denoise the corrupted labels during training. Similar to before, the proposed dropout scheme outperforms every scheme, including the true noise model, except for the 70\% noise level. However, dropout significantly outperforms bootstrapping in all regimes; at 70\% noise, dropout performs even better than bootstrap does at 50\% noise.

\begin{table*}[t]
\centering
\caption{Classification error rates for the DNN architecture trained on the MNIST dataset corrupted by uniform noise.}
\label{mniunideep}
\begin{tabular}{|c|c|c|c|c|}
\hline
Noise level & True noise & Base model & Dropout & Bootstrapping (\cite{reed2014training}) \\ \hline
30\%        & 2.46                                                       & 3.42       & 2.41            &                                                               2 \\ \hline
50\%        & 3.72                                                       & 23.4       & 3.63            &                                                               45 \\ \hline
70\%        & 7.59                                                       & 45.33      & 8.77           &                                                                N/A\\ \hline
\end{tabular}
\end{table*}

\begin{figure}[htb]
        \centering
        \begin{subfigure}[b]{0.5\textwidth}
                \centering
                \includegraphics[width=\textwidth]{True_70RandBW.jpg}
                 \caption{True noise}
                \label{}
        \end{subfigure}%
        \quad
         \begin{subfigure}[b]{0.5\textwidth}
                \centering
                \includegraphics[width=\textwidth]{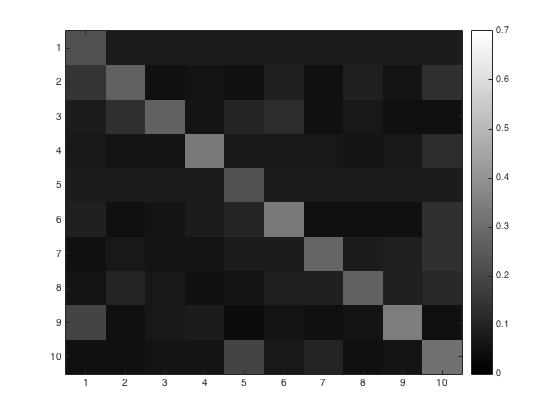}
                \caption{Learned noise}
                \label{}
        \end{subfigure}%
        \caption{True and learned noise model for the CNN architecture over the MNIST digits with 70\% label noise.}\label{fig:rand.70}
\end{figure}

Similar results obtain for non-uniform noise, as shown in Table \ref{mniranddeep}. Again, dropout has worse relative performance due to its difficulty in learning a non-uniform noise model, and this gap is significant at the 70\% noise level. We plot the true and learned noise model for the 70\% noise level in Figure \ref{fig:rand.70}. Similar to before, the learned model is more pessimistic and closer to a uniform distribution than the true model. We hypothesize that this has a more drastic effect because the MNIST digits do not cluster as naturally as the CIFAR images.

\begin{table*}[t]
\centering
\caption{Classification error rates for the DNN architecture trained on the MNIST dataset corrupted by non-uniform noise.}
\label{mniranddeep}
\begin{tabular}{|c|c|c|c|c|}
\hline
Noise level & True noise & Base model & Dropout & Bootstrapping (\cite{reed2014training}) \\ \hline
30\%        & 3.71                                                       & 6.03       & 2.45            &                                                               2 \\ \hline
50\%        & 5.24                                                       & 36.35       & 4.58            &                                                               45 \\ \hline
70\%        & 6.76                                                       & 53.55      & 43.03           &                                                                N/A\\ \hline
\end{tabular}
\end{table*}

While preparing this manuscript, we became aware of a recently-published approach \cite{azadi2015auxiliary}. It uses the ``AlexNet'' convolutional neural network, pretrained on a noise-free version of the ILSVRC2012 dataset. Then, for a different, noisy training set, it fine-tunes the last CNN layer using an auxiliary image regularization function, optimized via alternating direction method of multipliers (ADMM). The regularization encourages the model to identify and discard incorrectly-labeled images. This approach has a somewhat different setting---in particular, they rely on a pretrained CNN, whereas the results reported herein suppose that the end-to-end network must be trained via noisy labels---so we cannot give a direct comparison of our method to theirs. However, \cite{azadi2015auxiliary} reports a classification error rate of 7.83\% for 50\% noise on the MNIST set, whereas dropout achieves 2.83\%. This suggests that at least in some regimes dropout provides superior performance.

\section{Conclusion and Future Work}\label{sect:conclusion}
We have proposed a simple and effective method for learning a deep network from training data whose labels are corrupted by noise. We augmented a standard deep network with a softmax layer that models the label noise. To learn the classifier and the noise model jointly, we applied dropout regularization to the weights of the final softmax layer. On the CIFAR-10 and MNIST datasets, this approach achieves state-of-the-art performance, and in some cases it outperforms models in which the label noise statistics are known {\em a priori}.

A consistent feature of this approach is that it learns a noise model that overestimates the probability of a label flip. One way to interpret this result is that the deep network is encouraged to learn to cluster the data---rather than to classify it---to a greater extent than one would expect from the noise statistics. In other words, it is better to let deep networks cluster ambiguously-labeled data than to risk learning noisy labels. The details of this phenomenon---including which noise model is ``ideal'' for training an accurate network---is a topic for future research.

\section*{Acknowledgment}

This work is supported in part by the US National Science Foundation award to XWC (IIS-1554264)

%\section{Conclusion}
%In this paper, we have shown the way to efficiently use the convolutional neural network when the clean class labels are not present. We have shown the better classification accuracy of the proposed approach  over the current state-of-the-art methods as well as with the base convolutional neural network model. we have obtained results on many datasets ranging from small to large datasets and in different settings.  

\bibliographystyle{IEEEtran}

\bibliography{ref}

\end{document}